\documentclass[sigconf]{acmart}
\pdfoutput=1 

\usepackage{graphicx}
\usepackage{url}
\usepackage{booktabs}
\usepackage{balance}
\usepackage{subfigure}

\renewcommand\footnotetextcopyrightpermission[1]{} 
\begin{document}

%
\title{Classi-Fly: Inferring Aircraft Categories from Open Data using Machine Learning}
\author{Martin Strohmeier}
\author{Vincent Lenders}
\email{firstname.lastname@armasuisse.ch}
\affiliation{%
  \institution{
  armasuisse Science and Technology}
  \city{Thun}
  \country{Switzerland}}

\author{Matthew Smith}
\author{Ivan Martinovic}
\email{firstname.lastname@cs.ox.ac.uk}
\affiliation{
  \institution{Department of Computer Science, \\University of Oxford}
  \city{Oxford}
  \country{United Kingdom}
}

%
\begin{abstract}
In recent years, air traffic communication data has become easy to access, enabling novel research in many fields. Exploiting this new data source, a wide range of applications have emerged, from weather forecasting to stock market prediction, or the collection of information about military and government movements. Typically these applications require knowledge about the metadata of the aircraft, specifically its operator and the aircraft category.

\textit{armasuisse Science + Technology}, the R\&D agency for the Swiss Armed Forces, has been developing Classi-Fly, a novel approach to obtain metadata about aircraft based on their movement patterns.  We validate Classi-Fly using several hundred thousand flights collected through open source means, in conjunction with ground truth from publicly available aircraft registries containing more than two million aircraft. Classi-Fly obtains the correct aircraft category with an accuracy of over 88\%, demonstrating that it can improve the meta data necessary for applications working with air traffic communication. Finally, we show that it is feasible to automatically detect specific flights such as police and surveillance missions. 
\end{abstract}
%
%

\ccsdesc[500]{Information systems~Clustering and classification}
\ccsdesc[500]{Applied computing~Aerospace}
\ccsdesc[300]{Information systems~Business intelligence}

%
\keywords{aviation, aircraft classification, open datasets, air traffic control, air traffic management, open source intelligence}

%
\maketitle

\section{Introduction}

Crowdsourced aircraft trajectory data has gained significant importance in recent years enhancing data collection for several scientific fields and opening up new research opportunities. With the large-scale and open availability of globally distributed ground sensors, hundreds of research publications across different subjects have been exploiting this novel data source.\footnote{A regularly maintained list of examples can be found at \url{https://opensky-network.org/community/publications}.} Most, if not all, of these investigations require knowledge about the broad \textit{category} of an aircraft, e.g., whether it is a commercial airliner, a fighter jet, or a small private airplane; and its use case (e.g., government, business, surveillance, or military). 

This development has been enabled by the rise of software-defined radios (SDRs), which have become readily available and affordable over the past decade and greatly reduced the barriers to entry, meaning that users can now take part in crowdsourced sensor networks with little cost. As such, it is now straightforward to collect all air traffic surveillance communication directly from aircraft, providing information about flight trajectories, including its position, velocity and unique identifiers in order to tell them apart and track them over time. Based on this principle, many commercial ventures have been created, for example Flightradar24 and FlightAware, who in turn provide sophisticated data-based services to aviation stakeholders. 

In the wake of  COVID-19, the interest in trajectory-based research has accelerated even further. Relevant research can be separated into two different areas of modeling, \textit{pandemic} and \textit{economic}. 

The first area, epidemiological modeling of the possible spread of COVID-19, was of crucial interest early in the stages of the pandemic. The utility of flight data for this purpose was illustrated for example in widely circulated studies such as \cite{bogoch2020potential} but has been known to be useful in the context of pandemics for much longer (e.g., \cite{mao2015modeling}). Knowing aircraft category and operator can help improve the accuracy of these models, i.e. in estimating the number of passengers on each tracked aircraft.

The second main area, economic modeling, uses flights either as an indicator of economic activity (at a given airport, region, or globally) as illustrated in \cite{miller2020using} or as a direct measure of the impact on the aviation sector (in particular cargo and passenger transport). Here, the speed of these indicators is the crucial advantage, as they allow to `nowcast' the economy faster than traditional methods. Examples of such use of data provided by OpenSky can be found in the Bank of England's quarterly Monetary Policy Report \cite{bankmpc} or the National Statistics Office of Denmark \cite{UNdenmark}. Again, obtaining accurate metadata about the operators and purpose of the tracked aircraft is crucial for improving the accuracy of these models.

Crucially, contrary to the trajectories, there is no helpful information about the aircraft types, categories or operators broadcast by the aircraft themselves. To solve this issue, researchers rely broadly on external databases, maintained through a mix of crowdsourcing by aviation enthusiasts and official databases provided by a few countries' aviation agencies such as the US Federal Aviation Administration (FAA) \cite{FAA}. Unfortunately, these sources are of only limited use as they are incomplete and outdated for many of the world's aircraft, which poses a major challenge to applications in this area.

In this paper, we present a machine-learning approach to solve this problem called Classi-Fly. We first deal with the challenges of incomplete and unreliable ground truth by verifying the categories of our dataset manually. Using this dataset, we show that it is feasible to automatically classify aircraft into different operator categories based purely on their learned flight movement patterns. Contrary to other classification approaches, Classi-Fly works on features directly derived from the trajectories, which cannot be altered by the aircraft operator, a crucial advantage over classical methods.

In this paper, we make the following contributions:
\begin{itemize}
\item On a dataset of 182,325 flights, we show that it is feasible to automatically learn the category of a given aircraft with over 88\% accuracy based solely on its flight behaviour. We build a model based on features derived from this behaviour and compare the accuracy of four different classifiers.
\item Using this approach, we classify previously unknown aircraft into different categories, effectively deriving metadata information for these aircraft, which can be used for popular research applications from open-source intelligence to epidemiological modeling.
\item We discuss the implications of our method, including potential countermeasures, and analyze a case study of previously unidentified aircraft with sensitive mission profiles.
\end{itemize}

The remainder of this paper is structured as follows: Section \ref{sec:Background} describes the necessary background on air traffic control and tracking. Section \ref{sec:Classi-Fly} describes the fundamentals of our approach. \ref{sec:Data-Collection} introduces our data collection process. Section \ref{sec:Experimental-Design} describes our experimental design before Sections \ref{sec:Model} and \ref{sec:Unknown} present the performance of the analyzed classifiers on the ground truth and new data, respectively. Finally, Section \ref{sec:Discussion} and \ref{sec:Related-Work} cover the discussion and related work before Section \ref{sec:Conclusion} concludes.

\section{Background}\label{sec:Background}

\begin{figure}
\includegraphics[width=0.65\columnwidth]{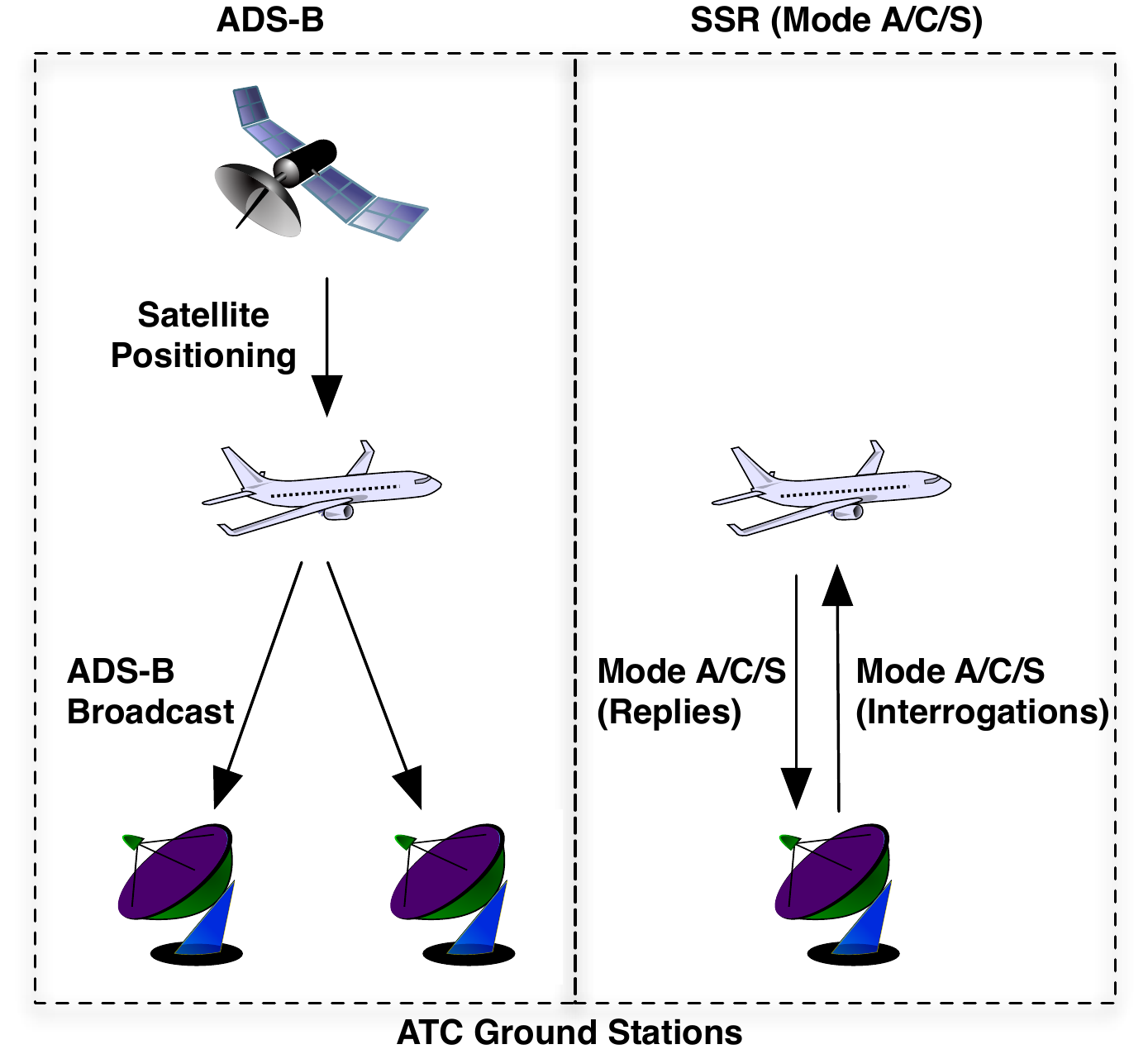}

\caption{Representation of ADS-B and SSR systems.\label{fig:Representation-of-ADS-B} }

\end{figure}

This section provides the necessary background to how aircraft tracking
works. Fig.\,\ref{fig:Representation-of-ADS-B} shows the wireless communication links of two considered technologies, which are explained in the following.

\subsection{Surveillance Technologies in Aviation}

There are two main surveillance technologies used for cooperative tracking of civil aircraft.
Secondary Surveillance Radar (SSR) uses the so-called transponder Modes A, C, and S, which
provide digital target information (altitude, squawk identification) compared to traditional
primary radar (PSR). Aircraft transponders are interrogated
on the 1030 MHz frequency and reply with the desired information on
the 1090 MHz channel (see Fig.\,\ref{fig:Representation-of-ADS-B}, right.) With the newer Automatic
Dependent Surveillance-Broadcast (ADS-B) protocol (see Fig.\,\ref{fig:Representation-of-ADS-B}, left),
aircraft regularly broadcast their own identity, position, velocity
and other information such as intent or emergency codes.
These broadcasts do not require interrogation; position and velocity
are automatically transmitted at 2\,Hz \cite{Schaefer14}.

\subsection{Aircraft Identifiers in Air Traffic Communication }\label{sec:identifiers}

A 24-bit address assigned by the International Civil Aviation Organization
(ICAO) to every aircraft is transmitted via both ADS-B and SSR. Crucially, this identifier
is different to an aircraft \textit{squawk} or \textit{call sign}. Squawks, of which
only 4096 exist, are allocated locally and not effective for continuous
tracking. The call sign can be set separately through the flight deck
for every flight. Call signs
of private aircraft typically consist of the aircraft registration
number, commercial airliners use the flight number, and military and
government operators often use special call signs depending on their
mission. In contrast, the ICAO identifier is globally unique and provides
an address space of 16 million; while the transponder
can be re-programmed by engineers, the identifier is not easily (or
legally) changed by a pilot. These characteristics make it
ideal for continuous tracking over a prolonged period of time.

\subsection{Required Data Mining Capabilities}

Aircraft tracking is the act of obtaining live or delayed
positional information on aircraft by passive actors. Their
motivations range from traditional hobbyist planespotting enthusiasm over military
and business interests to environmental science. Where traditionally most
spotters have conducted their trade purely using visual means, i.e., seeing and recognizing
the aircraft near an airport, modern software-defined radio (SDR) technology has made accurate, fast and scalable
tracking of aircraft feasible for anyone. 

There are two options to exploit SDRs: install their own personal receivers or use the SDR data aggregated by web tracking services. While a single receiver with a radius of up to 600 km can already
provide interesting results, the insights are increased considerably with
a larger network. Both live tracking data and
the required metadata are easily accessible on-line as discussed in Section \ref{sec:Data-Collection}.

\begin{table}[]
\centering
\caption{Description of the ground truth dataset, comprising 9880 randomly selected aircraft with a maximum of 25 flights. \label{tab:Description-truth}}

\begin{tabular}{@{}lrrr@{}}
\toprule
       & Flights / Aircraft & States / Flight & Duration / Flight [s] \\ \midrule
Mean    &   20.3148       &   152.62       &   4669      \\
Median &     25     		 &      79    	  &   1897      \\ \midrule
Total & 200,710         & 30,633,219       &    937,223,660     \\
\bottomrule
\end{tabular}

\end{table}

\section{Classi-Fly}\label{sec:Classi-Fly}

Classi-Fly is a machine-learning approach to categorize an aircraft purely on its behaviour, i.e. the way it moves over time rather than relying on self-reported information such as call signs and / or fallible databases. 

More concretely, Classi-Fly analyses collected trajectories of aircraft and breaks them down into 12 principal features, based on flight duration, position, velocity and acceleration. Using these features, it learns the behaviour of 8 different aircraft categories, from commercial to military and government aircraft. This in turn enables the user to automatically classify aircraft improve a knowledge base that is often incomplete and erroneous.

\subsection{Applications}

During the COVID-19 pandemic, aircraft trajectory-based research has accelerated significantly. Scientific modeling has used aircraft trajectory data to measure the impact of the implemented mitigation measures on the economy \cite{Iacus20,Garcia20}, nature \cite{Lecocq20} and air traffic itself \cite{Suzumura20}. The growing interest is reflected in the popularity of new datasets, which have seen significant uptake in a short time \cite{Olive20}.

Beyond this, journalists and scientific researchers alike have previously examined aircraft trajectories to better understand the security and privacy of the global air traffic system. For example, reporters have used such data to uncover federal surveillance aircraft \cite{buzzfeedspy}. More scientific use cases in the literature include the analysis of mergers and acquisitions data of public companies \cite{Strohmeier18b}, or the in-depth analysis of privacy leaks \cite{smith2018undermining}. 


To support these use cases, Classi-Fly can also contribute directly towards open data initiatives such as the OpenSky Network aircraft metadata database, which in turn is used for a wide variety of research applications (e.g., \cite{Strohmeier18b,smith2018undermining,schafer2017opensky}).

\subsection{Traditional Acquisition of Aircraft Categories}

The standard way to do obtain the required aircraft category uses the owner (or operator) and the aircraft model (e.g., corporate jets such as a Gulfstream), from which the use case and category of the aircraft may be inferred with good certainty. For instance, the Swiss Air Force operates aircraft for military operations, whereas business jets are likely used for corporate purposes. 

However, in a large percentage of cases there is no meta information
available for observed aircraft. This makes it much more difficult to identify the category of an aircraft. A recent study found that
around 15\% of all transponder-equipped aircraft could not be found
using publicly available data \cite{schafer2017opensky}. Typically, these aircraft are from countries that do not provide an open aircraft registry. Furthermore, they may be outdated entries or registered recently.


\subsection{Advantage of Behavioural Features}

In order to supplement and verify unreliable and incomplete metadata collected through external sources, Classi-Fly uses exclusively \textit{behavioural} features based on an aircraft's trajectories. The main advantage of this approach is that these features cannot be trivially altered or spoofed by the aircraft operator.  This is contrary to any classification based on the \textit{content} of their communication.

Non-behavioural features, which are based on content broadcast by an aircraft, are primarily its transponder code, the call sign and the squawk code, i.e., the aircraft identifiers described in Section \ref{sec:identifiers}. The latter two are set by the pilot (often in accordance with local customs and air traffic controllers) and can thus be adapted practically at will. The transponder code is also not reliable in many situations, e.g. when the US American Air Force One presented as a different, non-existent aircraft \cite{AF1}.

There is additional information about the capabilities of an aircraft provided by the Mode S Enhanced Surveillance (EHS) protocol features used by some aircraft. While interesting in theory, we have decided to not use these for our classification task for the following reasons: First, they, too, can easily be changed and manipulated by the aircraft operator at will. More crucially however, these communication options are not consistently used, over 50\% of aircraft do not broadcast any information besides position and velocity.

Requirements for Classi-Fly include robustness against malicious actors and intentional manipulation of the communication data by the aircraft operator. Consequently, all features that could be manipulated were excluded in our design. This leaves only behavioural features, which cannot be altered by an aircraft at all (e.g. increasing maximum possible velocity) or at least not without significant cost in terms of time and resources (e.g., diversions, distraction flights or other large changes to the mission pattern that make an aircraft look like a different class). Please note that while the content of ADS-B messages is fundamentally not secured, the behavioural features based on location and speed can be verified reliably using physical-layer methods such as multilateration \cite{strohmeier2017crowdsourcing}.

\section{Data Collection} \label{sec:Data-Collection}

We now describe the processes for the collection
of fine-grained tracking data and for obtaining aircraft ground truth
from public sources. All data used in this work has been openly available and is thus already accessible to researchers on an ever growing scale.

\subsection{The OpenSky Network}

OpenSky is a crowdsourced network \cite{Schaefer14}, which is used as the backbone
of our data collection. As of January 2020, the OpenSky Network
consists of more than 2500 registered sensors streaming
data to its servers. The network has currently received
and stored over 16 trillion ATC messages, adding over 20 billion messages
by more than 50,000 different aircraft every day. As a non-profit, research-oriented network, OpenSky offers open access to its data to academic researchers and has been used for a large number of publications spanning many different domains from aviation security to climate change research. Detailed information about the history, infrastructure and use cases of OpenSky are found in \cite{Schaefer14}.

\paragraph*{Data Acquisition and Pre-Processing}

Aircraft trajectories can be retrieved from the OpenSky Network for free for universities, flight authorities, and other non-profit research institutions.\footnote{\url{https://opensky-network.org/data/impala}} The available data goes back several years, for which it offers dense coverage of Europe and the USA. More recently, it has spread to other continents, although coverage in Africa in particular is still lacking as it is based on volunteers to provide the locally broadcast aircraft communication. We obtained about 200,000 such aircraft trajectories for our ground truth and another 180,000 for the different classification categories.

The raw data is obtained from OpenSky via an Impala shell and consists
of so-called \emph{state vectors}, which describe the state of every
observed aircraft, i.e., its position, altitude, and velocity in increments
of one second. All state vectors were then separated into
\emph{flights}, by dividing the
positional data messages received by all aircraft as follows: Each
positional state which is more than 10 minutes older than the next and is at an altitude of less than 2500\,m is considered an arrival state, and hence a finished flight. Note that not all flights seen by OpenSky are necessarily complete, if a flight begins or finishes outside the coverage area, the first/last message will constitute the end point of the flight. We did not differentiate between complete or incomplete flights in order to maximize the robustness of our approach. OpenSky conducts some additional processing to filter out erroneous messages and transmission-induced noise as well as potentially maliciously altered data \cite{schafer2018opensky}.

\subsection{Aircraft Behavioural Ground Truth} \label{sub:Aircraft-Behavioural-Ground}

To facilitate the feature selection in the next section, we required
ground truth on the average flight and movement behaviour of aircraft. We first retrieved the positional data of 9880 randomly selected aircraft seen by OpenSky
in the year 2017 to be able to obtain the average values as boundaries for our features. This data was capped at maximum of 25 flights per aircraft, which resulted in more than 200,000 collected flights, with an average duration of 4669 seconds and a total number of more than 30 million analyzed state vectors. Table \ref{tab:Description-truth} provides the details of the ground truth dataset.

We then used these randomly selected aircraft to learn the average
aircraft behaviour with regards to its flight features, which are
discussed in Section \ref{sub:Feature-Extraction}. For
each feature, we quantized the data set into $q$ quantiles and learned
these quantiles' specific bounds. These are then used to model the relative
behaviour of different aircraft categories for our classification task.

\subsection{Aircraft Metadata Ground Truth}\label{sub:Aircraft-Category-Ground}

There are several public sources which provide meta-information
on aircraft based on their identifiers: the aircraft registration
or a unique 24-bit address provided by ICAO. This typically includes
the aircraft model (e.g., Airbus A320) and the owner/operator (e.g.,
British Airways), which we exploited to label our aircraft category ground truth. 

We have used the following openly available sources to collect and verify the ground truth for our work:

\begin{itemize}
\item The OpenSky Network has recently released an aircraft database complimenting its tracking efforts with crowdsourced metadata on over 495,000 aircraft. Available here: \url{https://opensky-network.org/aircraft-database}
\item Another non-profit project, Airframes.org, is a valuable source, offering comprehensive metadata about 609,000 aircraft identifiers. This includes background knowledge such as pictures and historical ownership information (available at \url{http://airframes.org}).

\item For aircraft registered in the USA, the FAA provides a daily updated
database of all owner records, online and for download. These naturally
exclude any sensitive owner information but overall contain over 320,000
clean and well-organised records as of January 2018 (available at \url{https://registry.faa.gov/aircraftinquiry/}).

\item Furthermore, the plane spotting community actively maintains many
separate databases with spotted aircraft. They usually operate SSR
receivers and enrich the received data with information such as operator,
model, or registration manually. The database structure of Kinetic
Avionic\textquoteright s BaseStation software has become the de facto
standard format and is also used to exchange and share their databases
in forums and discussion boards. Our database version used stems from
November 2017, containing 455,457 rows of aircraft data. 

\item Lastly, web services such as FlightAware and Flight\-Radar24 provide online
access to more than a million aircraft IDs (available at \url{http://www.flightaware.com} and \url{http://www.flightradar24.com}).
\end{itemize}

When considering all these databases together, we had access to metadata for 2,180,803 unique aircraft identifiers; this snapshot for our work was taken in January 2018.

Note that these sources are naturally noisy, since they rely on compiling
many separate smaller databases, are often (partly) crowdsourced and change over time; aircraft are frequently registered, de-registered and transferred globally. Due to the number of aircraft involved in the experiments in this paper we could not verify the model and operator of every aircraft by hand (i.e., by following their behaviour on web trackers and ensure consistency with the existing database). Nonetheless, this is a realistic situation for anyone looking to accurately categorize aircraft and requires an approach which is robust to such noise fluctuations. 

\begin{table}
\caption{Description of the evaluation data set. \label{tab:Description-of-database-1}}
\begin{centering}
\begin{tabular}{@{}lrrrr@{}}
\toprule
A/C Category & Aircraft & Ratio [\%] & Flights & States [x1000] \\ \midrule
Business      & 1000      &    16.6        & 36,119	& 5196 \\
Commercial    & 1000      &     16.6       & 48,590	& 12,465\\
Fighter       & 921      &     15.3       & 6918	& 751 \\
Small Utility & 440      &     7.3       & 16,071	& 3360 \\
Surveillance  & 403      &     6.7       & 15,384	& 4571\\
Tanker        & 402      &     6.7       &	 7657 & 1125\\
Trainer        & 1080      &     18.0       &	 23,778 & 5602\\ 
Transport        & 768      &     12.8       &	 27,808 & 5067\\  \midrule
Sum           & 6014     & 100        & 182,325	&\ 38,142 \\ \bottomrule
\end{tabular}
\par\end{centering}

\end{table}

\subsection{Aircraft Category Extraction}\label{sec:categories}

Based on the trajectory data provided by OpenSky and the collected metadata, we obtained flight behaviour data for eight different aircraft categories described here in brief:

\begin{itemize}
\item \textbf{Business jets:} Business stakeholders typically fly jets capable of 4-20 passengers. Gulfstream's G-range, Cessna's Citation jets and Bombardier's Learjet and Challenger aircraft are amongst the most popular choices. However, this category also comprises smaller and larger aircraft as long as they are operated for business use.
\item \textbf{Commercial airliners:} A large group that makes up a vast majority of passenger miles in the air. It is defined by the operator, i.e. a commercial airline that conducts scheduled transport, typically with large aircraft seating 50 or more passengers (e.g., Airbus 320 or Boeing 737).
\item \textbf{Small utility aircraft (`general aviation'):} This aircraft group comprises a large variety of aircraft used privately and in commercial operations, which we class as so-called general aviation aircraft. Typical examples are the Cessna 172 and 182, the most sold aircraft models in the world. 
\item \textbf{Military fighter aircraft:} Fighters are designed primarily for air-to-air combat. Relatively few of these are equipped with ADS-B transponders; our group consists mainly of Eurofighters, Tornados and F15/16 aircraft.
\item \textbf{Military tanker aircraft:} These aircraft are capable of refuelling other aircraft in the air and provide essential operational capabilities. By far the most representative example in our dataset is the Boeing KC-135 Stratotanker. 
\item \textbf{Military trainer aircraft:} This category includes smaller jet and turboprop aircraft used as training vehicle for military pilots by air forces and navies around the world. Representative examples of such trainer aircraft are the Northrop T-38 Talon or the Pilatus PC-21.
\item \textbf{Military transport aircraft:} These are large aircraft used by the military to transport troops or equipment. Generally slower than aircraft intended for air fighting, they share some similarities with tanker aircraft. In our dataset these are represented mainly through the McDonnell Douglas/Boeing C-17 Globemaster III.
\item \textbf{Civil surveillance Aircraft:} These aircraft are used by police agencies for surveillance purposes. They are typically small utility aircraft with special equipment and exhibit particular behaviour during their missions.
\end{itemize}

We note that these categories are not determined solely on aircraft model but instead on their use cases as defined by the operator (i.e., military or not). Indeed, there is also overlap in some military aircraft models, for example Multi Role Tanker Transport (MRTT) aircraft fulfil several roles. 

Knowledge of these categories can help with a number of use cases. With the exception of the commercial airliners and small utility aircraft, all are directly potentially sensitive aircraft categories. Commercial airliners and business jets are required as input for research on economic activity (for example \cite{miller2020using}), while the latter are also particularly interesting for investment banking studies \cite{Strohmeier18b}. Civil surveillance aircraft as a category have played a role in uncovering clandestine operations by state and non-state-actors \cite{buzzfeedspy}, with small utility aircraft being the category that most of such surveillance aircraft masquerade as. In the military context, telling apart unidentified commercial aircraft from potential threats can make a difference in highly volatile situations such as the accidental shooting down of Ukraine International Airlines Flight 752 in Iran in January 2020 \cite{abintime}. Differentiating between the categories of fighters, tankers, trainers, and transport aircraft serves as an additional piece of information and can provide tactical advantages.

\begin{table}
\centering

\begin{minipage}{.45\columnwidth}
\caption{Top origin countries of the main dataset. \label{tab:known-origin}}
\centering
\begin{tabular}{@{}lrr@{}}
\toprule
Country & Aircraft & [\%] \\ \midrule
USA      & 2916      &  48.5          \\
Germany    & 816      &    13.6        \\
China       & 287      &      4.8      \\
UK  & 239      &   4.0         \\
Australia  & 212      &   3.5         \\
Netherlands        & 160      &      2.7      \\
Belgium        & 119      &    2.0        \\
Canada        & 110      &     1.8       \\ \bottomrule
\end{tabular}
\end{minipage}
\hfill
\begin{minipage}{.45\columnwidth}
  \centering
\caption{Top origin countries of unknown aircraft. \label{tab:unknown-origin}}
\begin{tabular}{@{}lrr@{}}
\toprule
Country & Aircraft & [\%] \\ \midrule
UK      & 121      &  11.4          \\
Austria    & 96      &    9.0        \\
Germany       & 71      &      6.6      \\
China  & 67      &   6.3         \\
Czech Rep.  & 59      &   5.5         \\
Ireland        & 53      &      5.0      \\
Australia        & 43      &    4.0        \\
Brazil        & 40      &     3.8       \\ \bottomrule
\end{tabular}
\end{minipage} 

\end{table} 

\section{Experimental Design} \label{sec:Experimental-Design}

We describe the features used to determine aircraft behaviour and explain the evaluation data set used to predict aircraft categories.

\subsection{Evaluation Data Sets}

To select our evaluation data set, we first queried the full sample of aircraft seen by OpenSky in January 2018, which spanned 87,000 aircraft in total. This sample was then classified into eight different categories based on operator and model metadata (see Section \ref{sec:categories}). 

We aimed to obtain 1000 aircraft per category, however, for five of the subcategories (in particular those comprising military and surveillance aircraft) there are fewer aircraft with reliable identification and the necessary transponder equipment required to obtain the detailed flight behaviour data. Thus, we picked all available aircraft for fighters, surveillance aircraft, tankers, trainer and transport aircraft.

For small utility aircraft, the available pool was larger, however, due to the fact that many surveillance aircraft share the same aircraft model (in particular Cessna 182's \cite{buzzfeedspy}), manual inspection of all aircraft and their tracks was required to accurately label the ground truth. For the abundant business and commercial categories, we picked random 1000 aircraft to represent their category.

Thus, the main data set used for our classification experiments consists of 6014 aircraft overall, each with a maximum of 50 flights. Table \ref{tab:Description-of-database-1} provides the breakdown of all aircraft categories as well as the number of flights and individual state vectors used to obtain the classification features. The lowest number of flights (6918) and messages (751,000) could be obtained for the 921 fighter aircraft, presumably due to their comparatively rare use. At the upper end, the 1000 commercial aircraft were seen on 48,590 flights with over 12 million messages, illustrating the high utilization of commercial airliners. Overall, more than 185,000 flights and almost 40 million messages were processed to obtain the behavioural features. Finally, Table \ref{tab:known-origin} shows the main countries of origin of our dataset, with the US making up just under half of all aircraft, followed by several European countries, China, Australia and Canada.

\begin{table*}[]
\centering
\
\caption{Description of features, based on quantization of each behavioural feature into $q$ parts.}
\label{tab:features}
\begin{tabular}{@{}lllr@{}}
\toprule
\textbf{ID}     & \textbf{Name}      & \textbf{Feature Description}                                                             & \textbf{Avg. RMI}   
                                                          \\ \midrule
                                                          \multicolumn{2}{l}{\textbf{Flight Level}}  & & \\ \midrule
$f_1,...,f_{q}$   & Duration           & Proportion of an aircraft's flight durations falling into $q$ quantiles.                    & 10.42\% \\
$f_{q+1},...,f_{2q}$  & Bounding Box       & Proportion of a aircraft's flight areas as bounded by a box falling into $q$ quantiles.  & 11.88\%   \\ \midrule

                                                          \multicolumn{2}{l}{\textbf{State Vector Level}} & & \\ \midrule
$f_{2q+1},...,f_{3q}$ & Altitude           & Proportion of altitude values recorded for the aircraft falling into $q$ quantiles.  & 11.84\%    \\ 
$f_{3q+1},...,f_{4q}$ & Heading            & Proportion of heading values recorded for the aircraft falling into $q$ quantiles.   & 8.66\%    \\ \midrule

$f_{4q+1},...,f_{5q}$ & X-Velocity           & Proportion of X-velocity values derived for the aircraft falling into $q$ quantiles.  & 16.63\%    \\
$f_{5q+1},...,f_{6q}$ & Y-Velocity           & Proportion of Y-velocity values derived for the aircraft falling into $q$ quantiles.   & 13.67\%   \\
$f_{6q+1},...,f_{7q}$ & Vertical Rate      & Proportion of vertical rate values recorded for the aircraft falling into $q$ quantiles. & 15.81\% \\
$f_{7q+1},...,f_{8q}$ & Heading Speed            & Proportion of heading speed values derived for the aircraft falling into $q$ quantiles.  & 13.57\%     \\ \midrule

$f_{8q+1},...,f_{9q}$ & X-Acceleration          & Proportion of X-acceleration values derived for the aircraft falling into $q$ quantiles.   & 14.53\%   \\
$f_{9q+1},...,f_{10q}$ & Y-Acceleration          & Proportion of Y-acceleration values derived for the aircraft falling into $q$ quantiles.   & 14.67\%   \\
$f_{10q+1},...,f_{11q}$ & Vertical Acc.    & Proportion of vertical acceleration values derived for the aircraft falling into $q$ quantiles. & 15.93\% \\
$f_{11q+1},...,f_{12q}$ & Heading Acc.           & Proportion of heading acceleration values derived for the aircraft falling into $q$ quantiles. & 16.07\%     \\

\bottomrule
\end{tabular}
\end{table*}

\paragraph*{Unknown Aircraft}

We further obtained all features described in Section \ref{sub:Feature-Extraction} from 1066 unknown aircraft, i.e., aircraft sending messages with identifiers where no metadata was available from any of the structured sources. We use the communication received from these identifiers to gain insights on the category of their aircraft. Naturally, we consider that there will be some noise in this dataset, which we will not be able to fully solve due to the lack of ground truth. Thanks to OpenSky's sanity checks, wrongly-received  identifiers caused e.g. by transmission or decoding errors have already been filtered out. 

Based on the 24-bit identifier, if truthful, it is possible to obtain the country the aircraft is nominally registered in, by comparing it with the official ranges defined by the ICAO \cite{icaoaddresses}. Table \ref{tab:unknown-origin} shows the main countries of origin, ranging from several European countries to China, Brazil and Australia. We find that the distribution is different to the main dataset (albeit with a small sample size), in particular the lack of US aircraft is noteworthy.

We have several hypotheses and explanations for the absence of these unknown aircraft from available public sources:

\begin{enumerate}
\item Sensitivity: Highly sensitive military or state aircraft are excluded from public records in most countries. Depending on their missions, their country, and their use cases, hobbyist plane spotters may not be able to fill these gaps with information gleaned through traditional planespotting.

\item Novel aircraft: Depending on the quality of the public or private records, aircraft in many countries take several weeks or months until they turn up in public databases.

\item No records available: Many countries' aviation authorities do not maintain a consistent and well-kept database in the first place. In others, such as Germany, privacy regulations are extremely strict, preventing aircraft records from finding their way into the public domain.

\item Wrong transponder ID: Finally, there are  occurrences, where the transponder ID setting of an aircraft does not match the public records, creating discrepancies in the metadata.
\end{enumerate}

\subsection{Feature Extraction}\label{sub:Feature-Extraction}

We selected 12 different features, divided into two categories:
flight level and state vector features. We explain these categories in the following; a full list of the chosen features is presented in Table \ref{tab:features}. 

\begin{figure}
\includegraphics[bb=0bp 0bp 1050bp 615bp,clip,width=1.385\columnwidth]{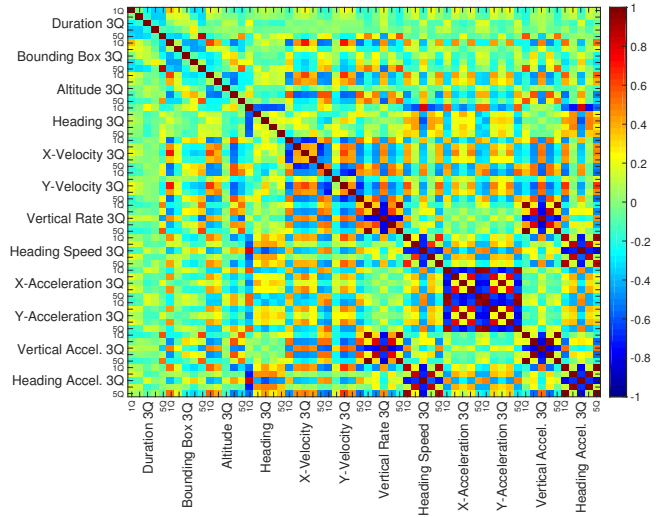}

\caption{Feature correlation matrix. 0 indicates no correlation,  1
and -1 positive and negative correlation, respectively. We can observe strong clusters of (anti-)correlations around both acceleration and velocity features.\label{fig:Feature-correlation-matrix}}

\end{figure}

\begin{figure*}
\includegraphics[bb=130bp 0bp 1030bp 470bp,clip,width=0.95\textwidth]{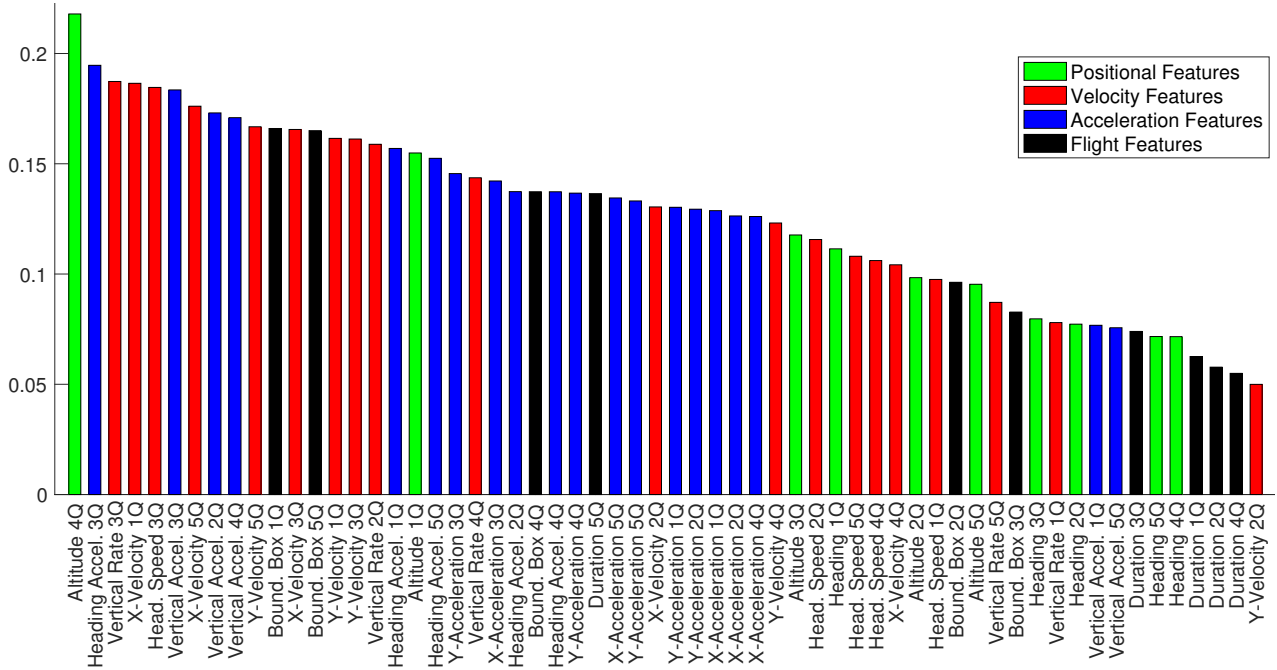}

\caption{Relative mutual information. Colors indicate the different physical feature groups. \label{fig:Feature-RMI}}

\end{figure*}

\subsubsection*{\textbf{Flight Level Features}}

These features contain information about the aircraft behaviour at
the highest level, namely the distribution of the \textit{durations} of all
its flights as well as the distribution of the \textit{area covered} by the
obtained flights of the aircraft. The distribution is represented
using the percentages of all flights falling into the chosen number of quantiles $q$
based on the average bounds obtained from the random sample in Section
\ref{sub:Aircraft-Behavioural-Ground}.

\subsubsection*{\textbf{State Vector Features}}

These features contain information at
the level of the collective state vectors, i.e., the distributions
of all of the aircraft's message content containing the heading, velocity,
vertical rate and altitude states. The distribution is based
on the training dataset as described in Section \ref{sub:Aircraft-Behavioural-Ground}
and represented as percentage of states falling into the chosen number of quantiles $q$.

There are three different types of state vector features based on their physical function: positional features, velocity features, and acceleration features (or the first and second derivative of the position with respect to time). Positional features include the altitude and heading values of their aircraft. The actual position in longitude and latitude values itself is not relevant, as it does not generalize to be a distinguishing feature across aircraft models and continents. Velocity features comprise the horizontal velocity in all three spatial dimensions as well as the speed with with the heading values of the aircraft change. Finally, acceleration features are derived with respect to time from all four of the velocity features.

\subsection{Feature Analysis}

\subsubsection*{\textbf{Feature Correlation}}

Fig.\,\ref{fig:Feature-correlation-matrix} shows the correlation between the features calculated on the evaluation dataset. An illustrative example is given by the heading quantiles. Here the first quantile (i.e., the ratio of no to few changes in aircraft direction) is strongly negatively correlated with all other heading quantiles. This suggests that many aircraft only ever have either few changes in directions such as commercial airlines, which stay in a straight line for most of their flight duration. Aircraft that have more or many directional changes in their flights can be clearly differentiated on this feature. Similarly, very high acceleration and deceleration are positively correlated in all three axes (X, Y, vertical), which reflects the capabilities and actual behavior of military fighter jets.

Beyond this, we can see strong relationships mainly between the horizontal velocity and acceleration features, aircraft with many values in high X-velocity and acceleration bins also exert this behaviour in the Y-direction. On the other hand, many aircraft either fall into long flights with constant middling speeds (e.g., commercial aircraft), or instead exert many very low and very high speed and acceleration values over the course of their flights, typical for fighter jets or trainer aircraft. On the other hand, few relationships can be observed on the flight durations.

\begin{figure*}
\includegraphics[bb=90bp 30bp 1420bp 810bp,clip,width=0.45\textwidth]{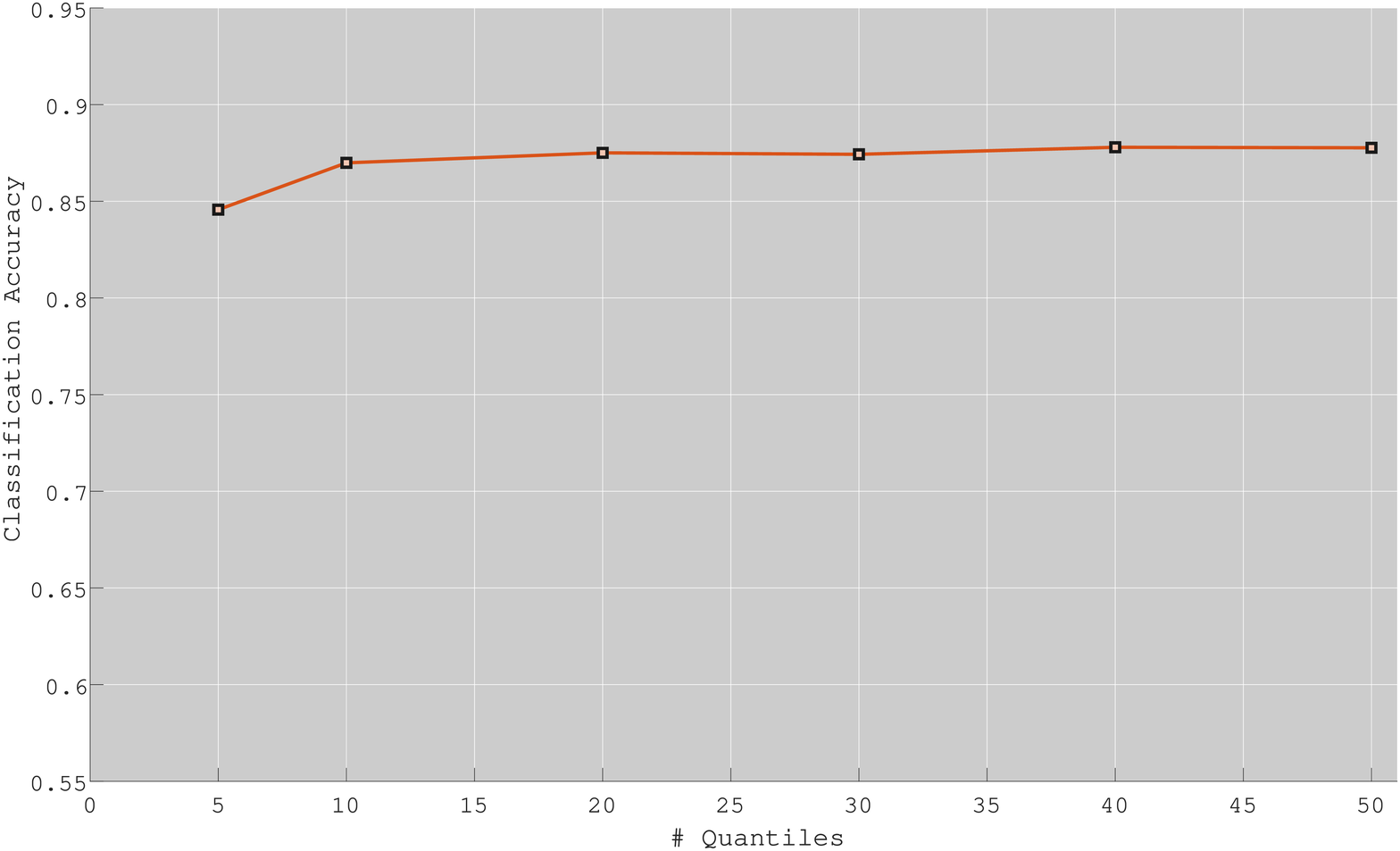}
\includegraphics[bb=90bp 30bp 1420bp 810bp,clip,width=0.45\textwidth]{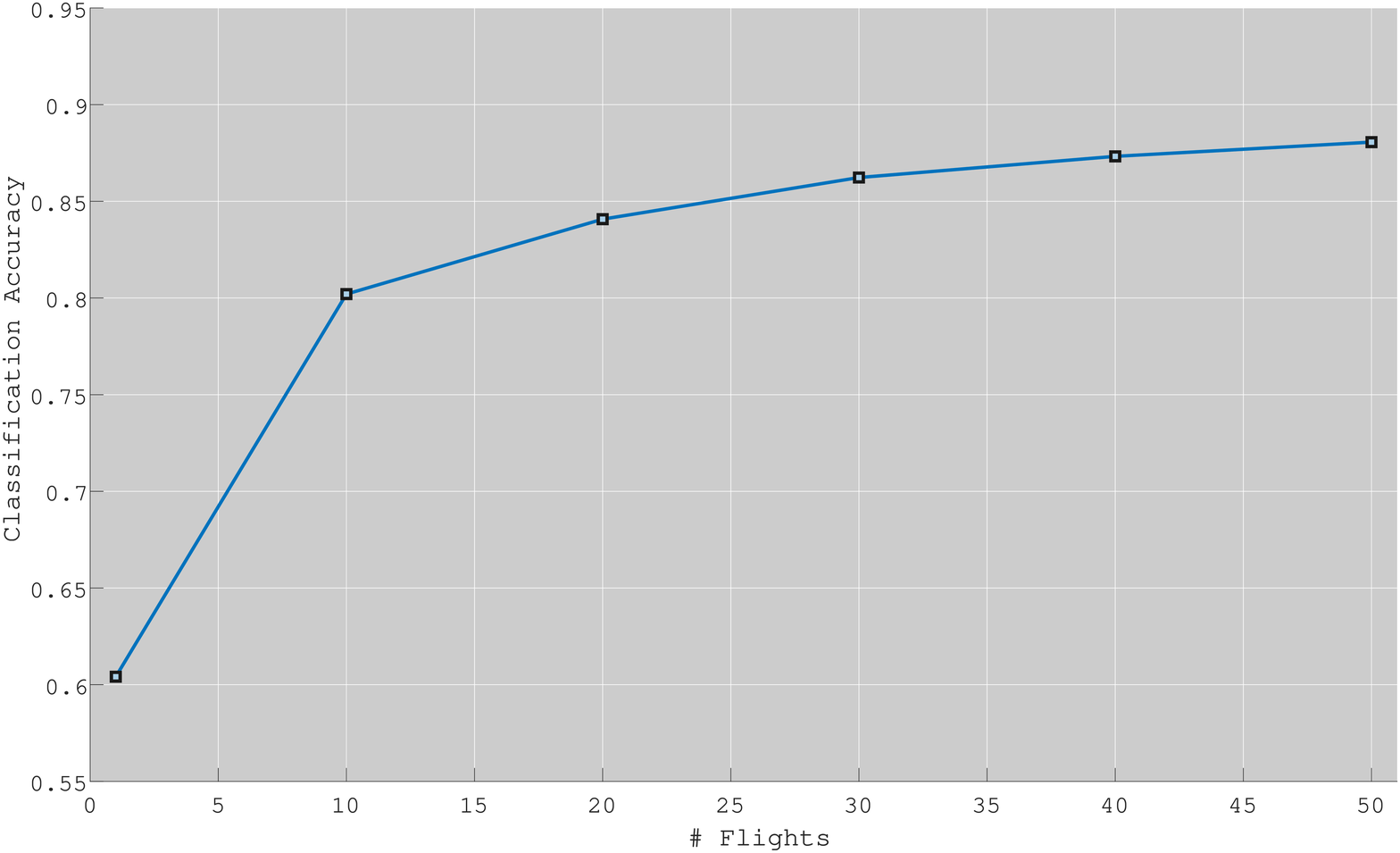}

\caption{Accuracy of the classification based on the number of flights $f_{min}$ (left) and feature quantiles $q$ (right).\label{fig:comparison}}

\end{figure*}

\subsubsection*{\textbf{Feature Quality}}

To obtain a clearer view on how the classification works and to identify potentially detracting features, we estimated their quality. There is a given amount of uncertainty associated with the aircraft category---its entropy. This amount depends both on the number of classes (i.e., aircraft categories) and the distribution of the
samples between them. As each feature reveals a certain amount
of information about the aircraft category, this amount can be measured
through the mutual information (MI). In order to measure the
mutual information relative to the entire amount of uncertainty,
the relative mutual information (RMI) is used. RMI measures
the percentage of entropy removed from the aircraft category ($cat$)
when a feature ($F$) is known \cite{battiti1994using}. 

The RMI is defined as

\begin{equation}
RMI(\textit{cat},F) = \frac{H(\textit{cat}) - H(\textit{cat}|F)}{H(\textit{cat})}
\end{equation}

where $H(A)$ is the entropy of $A$ and $H(A|B)$ denotes the entropy
of $A$ conditional on $B$. In order to calculate the entropy of a
feature it has to be discrete. As most features are continuous
we perform discretization using an Equal Width Discretization
(EWD) algorithm with 20 bins \cite{dougherty1995supervised}. This algorithm typically
produces good results without requiring supervision.  As outliers may have a drastic effect on the RMI computation, we use the $1^{st}$ and $99^{th}$ percentile instead of the minimal and maximum values to compute the bin boundaries in order to prevent large distortions. A high RMI indicates that the feature is distinctive on its own, but
it is important to consider the correlation between features as
well when choosing a feature set. Additionally, features may be more distinctive when combined, even when they are not particularly useful on their own.

Figure \ref{fig:Feature-RMI}  shows the RMI for each of our selected behavioural features, the colors indicating their physical feature group (positional, velocity, acceleration, or flight level). Overall, the velocity and acceleration features (red and blue, respectively) share the most information with the aircraft category, with many of these having an RMI of 15\% or more. The positional and flight level features 
are relatively less distinctive, which suggests that for example the distribution of heading values or the overall flight durations are more common to any aircraft mission than a consistent behavioural feature of a category. However, we choose to keep all features for our classification to produce the best results.

\subsection{Effects of Number of Flights and Feature Quantiles}

Finally, we more closely examined the effects of two feature parameters on the accuracy of the classification: the number of flights $f_{min}$ collected for each aircraft's feature creation, and the number of quantiles $q$, into which the state vector features were divided. Fig.\,\ref{fig:comparison} illustrates these relationships, by training a Support Vector Machine with varying values of $f_{min}$ and $q$. 

The minimum number of flights required to be create a feature vector has a significant effect on classification accuracy. With no lower bound, the overall classification accuracy is fairly poor at 61\%. Such a result is likely due to the classifier accounting for lots of edge cases, making it less generalized. Performance quickly increases to over 80\% with 5 collected flights; increasing the number of flights per aircraft further, the accuracy increases to over 85\% at 30 flights and 88.1\% at 50 flights. However, by raising $f_{min}$ to 50, the training set size decreases substantially---we found an $f_{min}$ of 30 to be a reasonable balance between data set size and accuracy. All results were obtained with $q = 10$ and represent the mean of 100 classifications.

The number of feature quantiles is also related to classification accuracy. Intuitively, as the number of quantiles increases, the risk of overfitting may increase. With the minimum of $q = 5$ the accuracy was 84\%, increasing to 87\% at $q = 10$, and only increasing marginally thereafter until leveling off at $q = 40$ and 87.8\%.  Further increases to $q = 50$ show no positive effect. As such, we found 10 quantiles to be a good balance of accuracy and generality. For the analysis, $f_{min}$ of 30 was used, with scores averaged over 100 repetitions.

\begin{table}
  \footnotesize
  \caption{Summary of model optimization and training results across four types of classifier.}
  \begin{tabular}{@{}llll@{}}
    \toprule
    Model Type & Parameters & Avg. Accuracy & Avg. TP Rate \\ \midrule
    \begin{tabular}[c]{@{}l@{}}Support\\ Vector\\ Machine\end{tabular} & \begin{tabular}[c]{@{}l@{}}Kernel Function: Cubic\\ C=4.795\\ Multiclass Method: One vs. All\\ Standardized Input\end{tabular} & 86.0\% & 84.4\% \\ \midrule
    \begin{tabular}[c]{@{}l@{}}K--Nearest\\ Neighbors\end{tabular} & \begin{tabular}[c]{@{}l@{}}Distance Metric: City Block\\ Distance Weight: Inverse Weight\\ \# Neighbors: 4\end{tabular} & 84.6\% & 83.9\% \\ \midrule
    \begin{tabular}[c]{@{}l@{}}Decision\\ Tree\end{tabular} & \begin{tabular}[c]{@{}l@{}}Criterion: Gini Index\\ Max Splits: 1297\end{tabular} & 71.4\% & 67.9\% \\ \midrule
    \begin{tabular}[c]{@{}l@{}}Ensemble\\ (Random\\ Forest)\end{tabular} & \begin{tabular}[c]{@{}l@{}}Method: AdaBoost Decision Tree\\ \# Learners: 402\\ Max Splits: 125\\ Learning Rate: 0.792\end{tabular} & 85.9\% & 84.0\% \\ \bottomrule
    \end{tabular}
    \label{tab:class-compare}
\end{table}

\begin{table}[]
  
  \caption{Summary of metrics for evaluation runs of each classifier. Metrics are averaged across scores from each class.}
  \begin{tabular}{@{}lllll@{}}
    \toprule
    & Accuracy & Precision & TP Rate & TN Rate \\ \midrule
   SVM & 85.3\% & 82.6\% & 84.4\% & 97.5\% \\
   KNN & 84.2\% & 81.1\% & 84.3\% & 97.2\% \\
   Decision Tree & 71.4\% & 67.9\% & 65.1\% & 94.4\% \\
   Random Forest & 86.7\% & 82.5\% & 86.1\% & 97.7\% \\ \bottomrule
  \end{tabular}
  \label{tab:eval}
\end{table}

\begin{figure*}
  \centering
\subfigure[SVM Evaluation Confusion Matrix]{\includegraphics[width=.8\columnwidth]{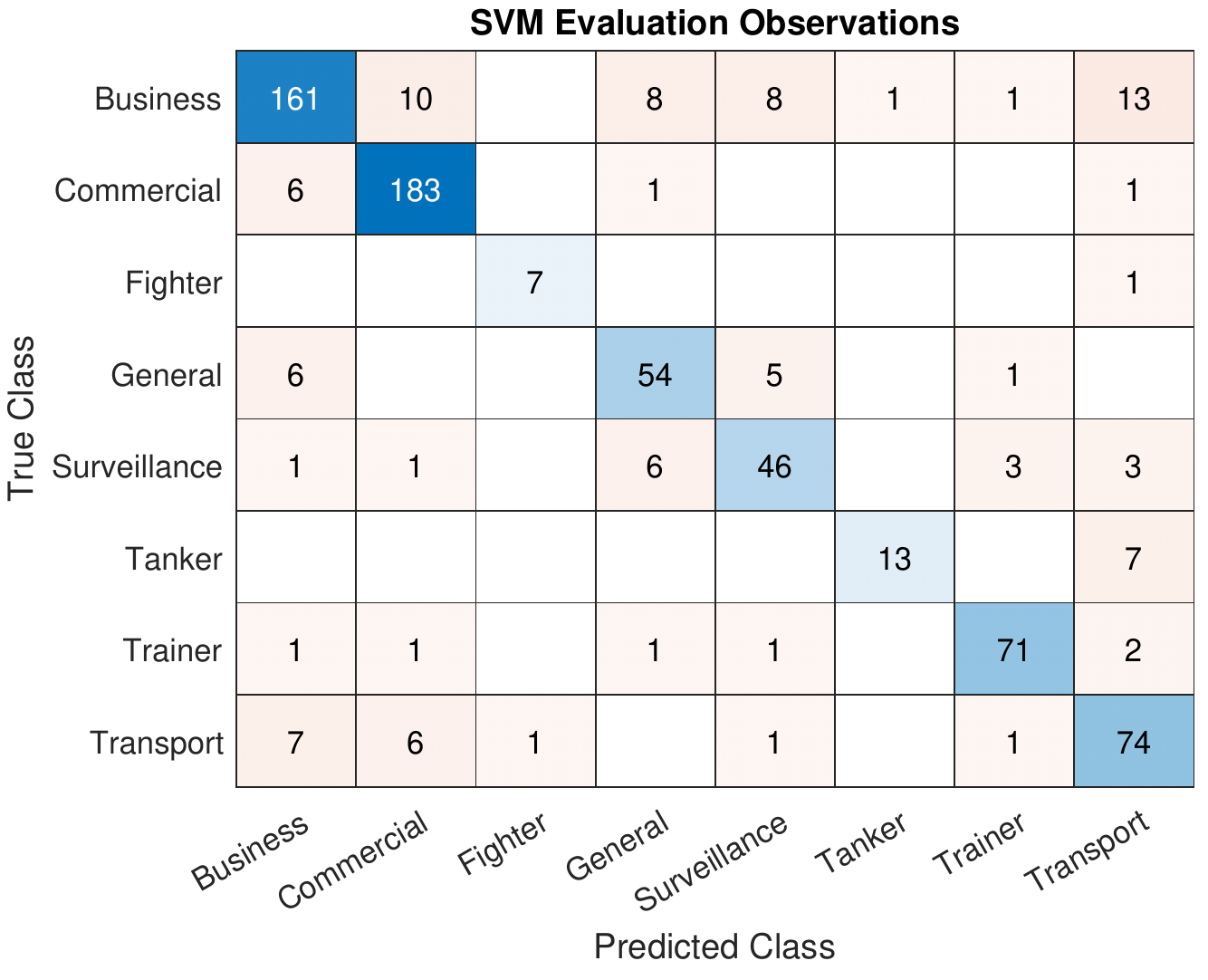}}
\subfigure[KNN Evaluation Confusion Matrix]{\includegraphics[width=.8\columnwidth]{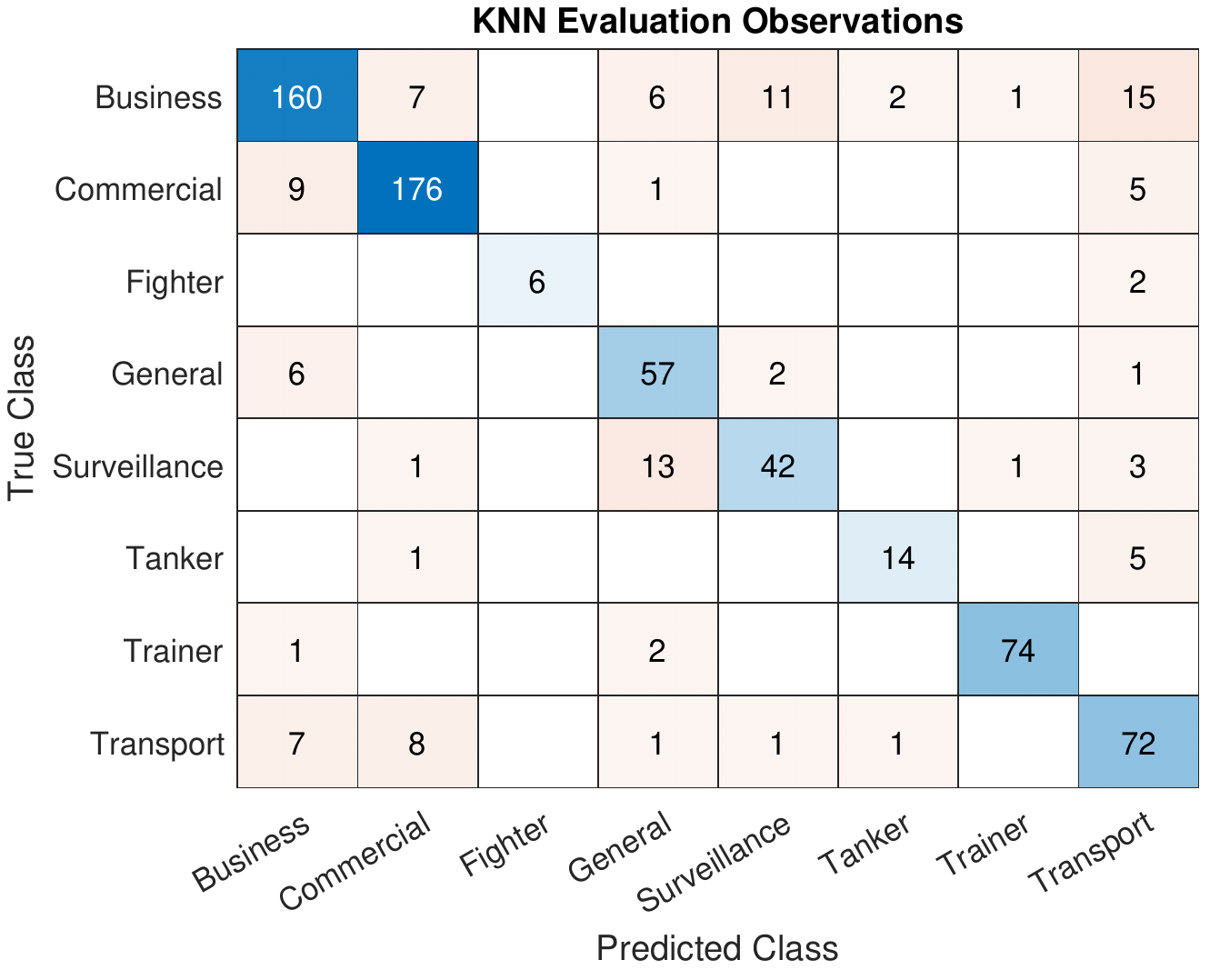}}\\
\subfigure[RF Evaluation Confusion Matrix]{\includegraphics[width=.8\columnwidth]{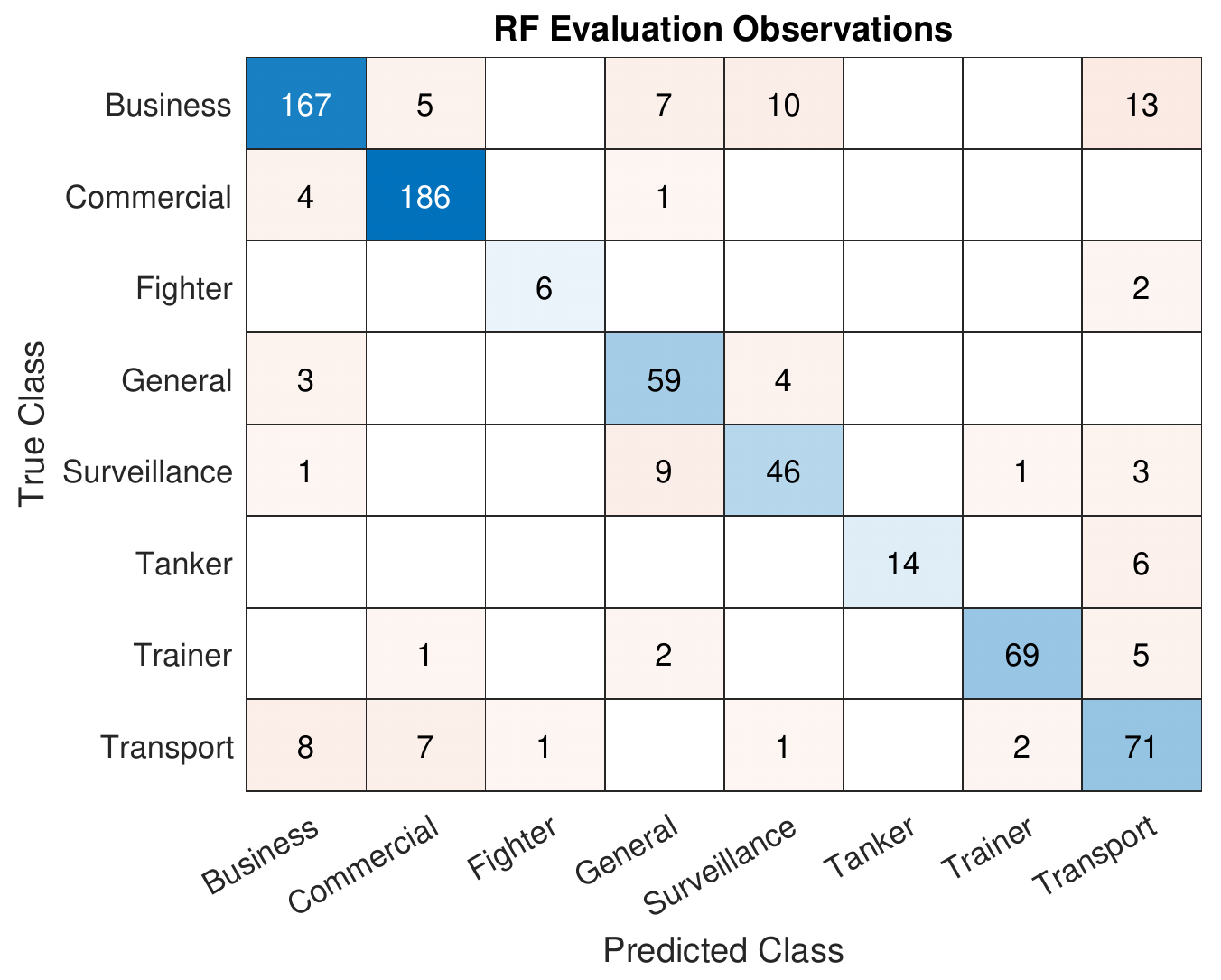}}

\caption{Confusion Matrices of observations on the evaluation set for the Support Vector Machine, Random Forest and K-Nearest Neighbors classifiers.}
\label{fig:confmats}
\end{figure*}

\section{Model Selection}\label{sec:Model}


We now compare the results of four classifiers for this classification task, implemented in Matlab: Decision Tree, Random Forests, Support Vector Machines and K-Nearest Neighbors. With the settings obtained in the previous section for $f_{min}$, we retained 3519 aircraft. We trained each classifier on 80\% of our training set (2859 instances), retaining 20\% (713 instances) for evaluation. We used 5-fold cross-validation on our training set to reduce overfitting.

\subsection{Classifier Training and Optimization}

The results of the classification show whether aircraft categories can be distinguished purely on their movement behaviour.  Based on our analysis above, we used a minimum number of flights $f_{min} = 30$ and number of feature quantiles $q = 10$. We trained a range of model types to allow for comparison: Support Vector Machines (SVM), K-Nearest Neighbors (KNN), Decision Tree and Ensemble methods (here Random Forests, RF). To find reasonable parameters for the models, we carried out a randomized search for 30 iterations.

As shown in Table~\ref{tab:class-compare}, SVM, KNN and RF offer similar performance during training across both accuracy and true positive rate. SVM performs slightly better across both metrics. 

Decision tree is the worst performing model of the four; it performed very well in identifying commercial and training aircraft but much worse in all other categories. This could be due to the limited ability of the model to handle labels with fewer training instances, with the tree instead mainly able to classify the more frequent labels.

\subsection{Classifier Evaluation}

After training, we evaluated our classifier performance on the unseen test portion of our dataset. As shown in Tab.~\ref{tab:eval}, performance is in line with performance during training in Tab.~\ref{tab:class-compare}. As before modelling with a decision tree offered poor performance relative to the other approaches, having a high TPR for the three most populous classes and relatively low TPR elsewhere. As such we do not consider this classifier further.

The remaining three classifiers have similar performance across the metrics in Tab.~\ref{tab:eval}. In Fig.~\ref{fig:confmats} we can see the number of observations predicted correctly by the SVM, KNN and RF models. As indicated by the performance metrics, each classifier performs similarly with slight biases towards certain classes. More specifically, KNN made more errors when classifying commercial aircraft than the SVM and RF models but performed better on the trainer and transport classes than the RF. 

All classifiers made similar classification errors. Transport aircraft were particularly susceptible to this, with transport often being predicted when the aircraft was actually business, or actual transport aircraft being misclassified. This is likely due to the low number of transport aircraft instances as well as their similarity to other aircraft movements or multi-purpose nature, i.e flying one-off or irregularly timed routes between special-purpose airports~\cite{gibson2002death}.

Of the three classifiers, RF has the lowest `spread' of misclassification, i.e. misclassified business aircraft fell into four categories rather than across all the other seven. This suggests the model is better generalized than the others, which have quite a few cases of single instance misclassifications in unusual places, e.g. SVM and KNN classifying a surveillance aircraft as commercial.

These errors also highlight the distinctiveness of some aircraft compared to others. We can see that trainer aircraft are rarely misclassified and when they are, the predicted class falls into one of four other labels. Business aircraft, however, have misclassifications across a range of labels. This could be a result of business aircraft being used for a wide range of purposes which in turn might result in flights similar to other categories.

If a single classifier is needed, the RF or SVM models provide equally good across-the-board performance in comparison to KNN. However, any of the three models would perform quite well and could be used to construct a meta-classifier. Further training examples in the lower population classes would help to explore whether certain classifiers perform better for some classes than others, helping to better assess the benefit of a meta-classifier. However, this would need to be done with care as individual classifiers good at identifying certain types of aircraft might be outvoted.

\section{Analysis of Unknown Aircraft}\label{sec:Unknown}

Table \ref{tab:unknown-aircraft} shows the classification of approximately 1000 aircraft, about which there was no data available in any publicly accessible database at the time of our snapshot. All selected aircraft had at least 10 flights and 500 state vector data points available for their feature creation, to reduce the amount of noise to a minimum and ensure that these are consistently used aircraft identifiers. To obtain categories for these aircraft, we used the random forest classifier trained on the known aircraft data as described above. As an ensemble classifier it provides confidence scores, i.e., the percentage of times a sample has been classified as a particular category. We used these scores as a cut off threshold, i.e., any sample classified with a score of less than 0.5 in any of the eight classes was judged as too low to provide useful insights. Taking this into account, 52.3\% of all aircraft were classified confidently into one group. Table \ref{tab:unknown-aircraft} shows the full results.

The commercial aircraft could overwhelmingly be verified manually using the most current online source, FlightRadar24, as having been put into service after the time our metadata snapshot was taken in January 2018. Indeed, of the 316 aircraft, 305 were classified correctly, with the 11 misclassifications being larger business jets. The new airliners in this set included, for example, 9 Boeing Dreamliners delivered to Norwegian in the first half of 2018 \cite{norwegian} or new aircraft in China, the biggest growth market for commercial aviation.

We further find that a large number of aircraft are seen by the classifier as business and small utility aircraft (10.9\% and 4.6\% respectively). This is plausible, as information on such private aircraft is not necessarily well-publicized, potentially even sensitive, and many countries other than the English-speaking world either do not require such aircraft to be on a public register or even do not publish any aircraft register at all. While we can naturally still not verify the accuracy of the classification, many such classified aircraft are regulars at typical business airports (e.g., Farnborough, UK or Teterboro, US), improving our confidence. The final large group was made up of surveillance aircraft (6.9\%), whose sensitivity provides a clear motivation for not publishing their meta information. We discuss a detailed case study on such an aircraft in the next section.  There was a small minority of aircraft classified as trainer aircraft (0.2\%). Finally, no military fighter, transport, or tanker aircraft were found in this dataset.

\begin{table}
\centering

\caption{Classification of unknown aircraft. \label{tab:unknown-aircraft}}

\begin{tabular}{@{}lll@{}}
\toprule
Aircraft Category & Aircraft & Percentage \\ \midrule
Business      & 116      &  10.9\%          \\
Commercial    & 316      &    29.6\%        \\
Fighter       & -      &      -      \\
Small Utility  & 49      &   4.6\%         \\
Surveillance  & 74      &   6.9\%         \\
Tanker        & -      &      -      \\
Trainer        & 2      &    0.2\%        \\
Transport        & -      &     -       \\ \midrule
Other        & 509      &   47.7\%         \\ \midrule
Sum           & 1066     & 100 \%       \\ \bottomrule
\end{tabular}
\end{table}

\subsubsection*{Detection of Surveillance Aircraft}

\begin{figure}
\includegraphics[width=0.865\columnwidth]{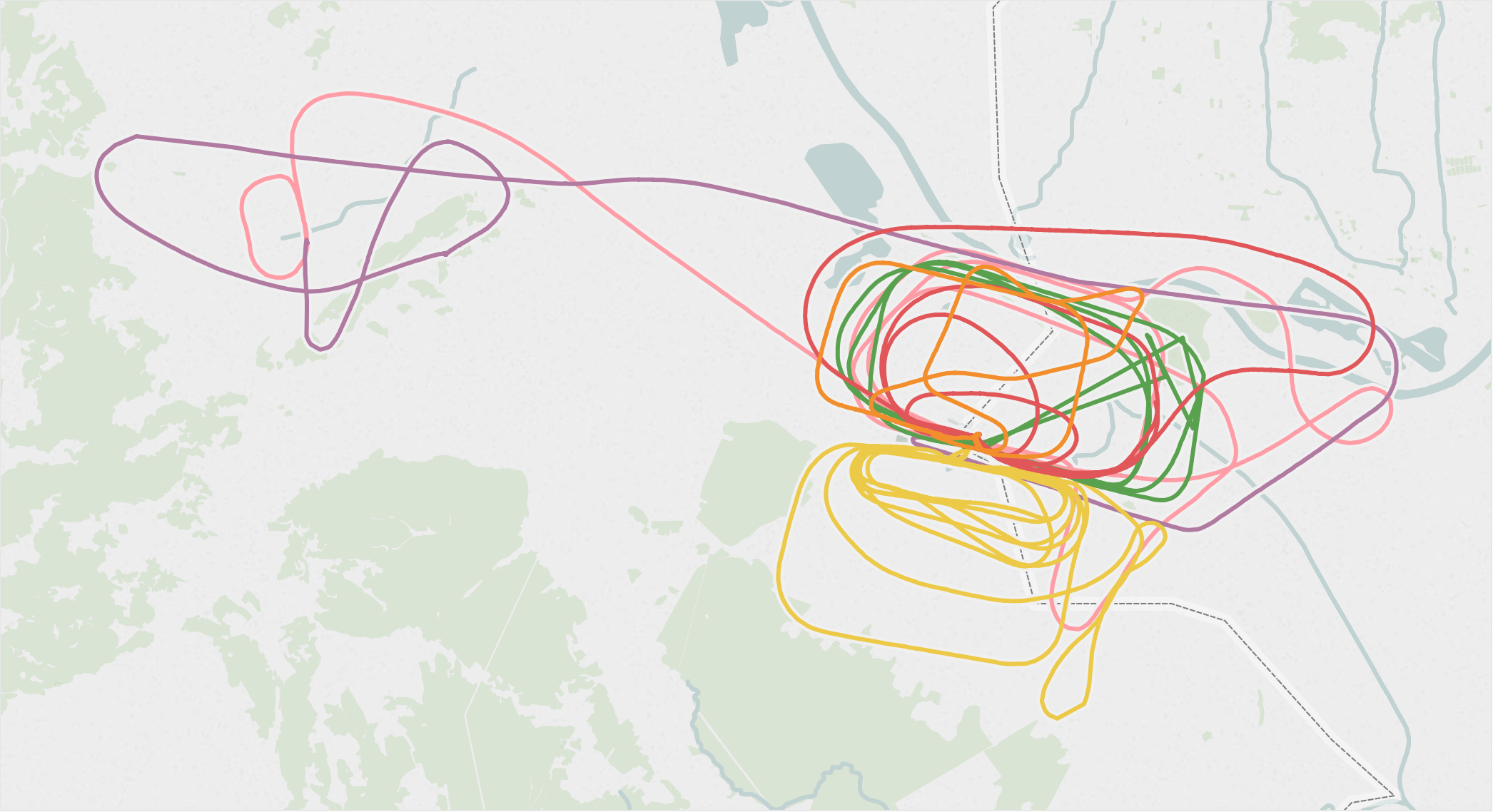}

\caption{Example of seven flight trajectories from a previously unknown surveillance aircraft detected in Croatia. Each colour is an individual flight and all flights clearly exhibit `'circling' features exhibited by surveillance aircraft.\label{fig:spy-example}}
\vspace{-5pt}
\end{figure}

We now take an example case study of `interesting' aircraft categories being detected by Classi-Fly. As first laid out in \cite{buzzfeedspy} for the case of the United States, there are many federal surveillance operations conducted with undercover aircraft. Such aircraft register with inconspicuous call signs, registrations and transponder identifiers, thus evading all database-based detection and making behavioural classification necessary. In the case of surveillance aircraft, some of the defining features can be visualized well just with the basic trajectory data. Most notably, such aircraft do not fly from point A to point B in a relatively straightforward fashion. Instead, they fly to a target area (say, a border area or an ongoing high-profile criminal incident) where they circle steadily in search of persons or objects, shifting altitude and center position occasionally.

Fig. \ref{fig:spy-example} visualizes this precise behavior. It shows seven flights from an example classified with very high confidence (RF score of 0.91) as surveillance aircraft.  While no information about this aircraft is available, as it does not appear in any database, it clearly exhibits the patterns of an aircraft used for surveillance of a narrow area, which are picked up by the classifier. The number of flights with this nature clearly increases confidence in the correctness of this prediction. On further analysis, we found that these flights were conducted in Croatia and we speculate that they are related to military and anti-terrorist missions.

This case study shows that our approach generalizes across different countries and their surveillance institutions and is able to detect surveillance aircraft around the globe.

\section{Discussion}\label{sec:Discussion}

We now discuss the limitations of Classi-Fly, possible refinements, and potential countermeasures to our approach.

\subsection{Limitations}
The greatest limitation of Classi-Fly is the inherent non-specificity of some categories. For example, it is difficult to identify the precise use case of a business aircraft; besides business travel, the same Gulfstream G550 could be used for transport of goods for the military or people for leisure, which can pose a potential threat to the validity of the results. Likewise, there may be differences between countries and institutions not captured in these categories and the ground truth. As long these do not pertain to behaviour, our methods work well but it is conceivable that different global regions or military institutions may also exhibit differences in behaviour with their aviation hardware.

However, with further research into potential subcategories and how to define them based on metadata such as the operator or owner or the airports frequented, this could be mitigated and their different behavior learned. This applies also to currently neglected aircraft categories such as unmanned aerial vehicles (UAV, or drones) and ultralight aircraft (ULAC), which will become transponder-equipped in larger numbers in the future and are a major interest for researchers and the aviation industry.

\subsection{Refinement}

Besides improving the ground truth metadata on the examined aircraft categories, other, non-behavioral features can be integrated into Classi-Fly. As many wireless standards (not only in aviation) give manufacturers a large amount of freedom over the actual soft- and hardware implementations, differences emerge that can be used as classification features.

On the physical layer, \cite{leonardi2017air} proves that it is feasible to distinguish aircraft transponders based
on anomalies in the frequency stability of their messages. On the data link layer, research has exploited differences in the transponders' random backoff algorithms \cite{strohmeier2015passive}. 

Besides these approaches, it is possible to add a host of features derived from the actual message content sent out by the aircraft. In a non-adversarial setting where the aircraft operators do not actively seek to obfuscate their identity (beyond excluding it from public databases), this would greatly improve classification accuracy.

Overall, we assume that certain uncommon aircraft may be
individually identifiable through the combination of features. Future work will thus consider the possible granularity that several approaches can provide if they are combined and further quantify the privacy impact for aircraft owners and operators.

\subsection{Countermeasures}

As our classification approach is agnostic to any non-behavioral
features, it is more difficult to apply effective countermeasures against
it.  Aircraft could deliberately change their behavior to avoid
detection and classification. However, this forces
the aircraft into not being able to fulfil its intended function freely, for example surveillance aircraft not circling their target, or military fighter jets deliberately flying slowly. This limits the potential benefit of such an option.

Related work \cite{Strohmeier18b} has looked at countermeasures
to the basic enabling mechanisms of aircraft tracking, which is
generally based on the ICAO identifier or other directly identifying information broadcast
voluntarily by the aircraft (such as its registration). There are
two popular privacy-preserving approaches to aircraft tracking found
in the aviation industry: the first consists of not displaying aircraft
on popular web feeds (such as FlightRadar24 or FlightAware), the second
comprises the use of shell companies to hide the real owners of an
aircraft and thus undermine the collection of accurate metadata. Both
ideas, while certainly popular, are ultimately not effective against
a moderate threat model that allows for the independent localization of aircraft \cite{Strohmeier18b}. 

The most effective countermeasure as concluded by the literature consists
of the randomisation of the aircraft's ICAO identifier, making it
difficult to continuously track the same aircraft over time. If done
globally for all aircraft, and in conjunction with other pseudonymisation
measures regarding the registration, it could effectively thwart consistent
aircraft tracking and by extension also Classi-Fly. However, the cat
may largely be out of the bag already; with the current widespread availability
of comprehensive aviation data there is sufficient input available
for training.

\section{Related Work} \label{sec:Related-Work}

The classification of objects or subjects based on wireless communication has been a popular field of research, in particular with a focus on security and privacy aspects. Exemplary studies outside the aviation domain range from the mobility states of humans \cite{mun2008parsimonious} to the classification of intruders (people, soldiers, vehicles) in a military setting \cite{arora2004line}. 

The closest related academic research is the classification of different types of ground vehicles. Vehicle type classification is an important signal processing task with widespread military and civilian applications in  intelligent transportation systems \cite{duarte2004vehicle}. Several data types have been used for vehicle classifications, collected for example from acoustic or seismic \cite{wu1998vehicle,scholl1999seismic} sensor sources. While these may be applicable in the aviation domain, too, our work focuses on the trajectories for classification.
 
Regarding such trajectories, the authors in \cite{liu2010uncovering} used GPS-based tracks of cab drivers to study their behavior and classify them into high-earning and average-earning drivers through the use of angularity and travel time features. Using taxi tracks with a different focus, further work attempted to uncover anomalous trajectories in a dataset by comparing and isolating tracks which are few and different from the majority \cite{zhang2011ibat}.

Most closely related in the vehicle domain, the authors in \cite{sun2013vehicle} distinguish two classes of vehicles (trucks and passenger cars) using GPS data extracted from mobile traffic sensors with a misclassification rate of 4.6\%. The main features are based on the vehicles' acceleration and deceleration behavior. Our work transfers this idea into three dimensions and applies it to the very different speeds and vehicle types found in aviation.

In the aircraft domain, wireless classification has focused on traditional non-cooperative PSR communication as the medium. Such work exists for both military \cite{lin1981optimum} and commercial aircraft \cite{zyweck1996radar} and exploits for example Doppler signatures \cite{bullard1991pulse} and high resolution range profiles \cite{zyweck1996radar} to identify the type of aircraft seen by the radar. However, primary radars are prohibitively expensive and thus widely inaccessible for research. As they are being replaced globally with the more accurate and cost-efficient ADS-B, we choose to focus on this cheap and openly available source of aircraft trajectories.

Finally, the closest non-academic work related to our approach is the successful attempt of investigative journalists to uncover unknown surveillance aircraft in the USA, which was presented at DEFCON 25 \cite{defconskies}. The authors report on the background of so-called spy aircraft, which are identified using a machine learning approach on aircraft flight data pre-processed by a large commercial tracking website. While we follow a similar basic approach concerning such surveillance aircraft in this work, we systematically analyze the effectiveness and validity of applying machine learning to aircraft behavior. In order to do this, we  process a large open data set, and discuss requirements on features and number of flights. Furthermore, we generalize this approach to many aircraft categories. 

\section{Conclusion}\label{sec:Conclusion}

In this work, we presented Classi-Fly, a machine learning method to infer the categories of aircraft, both anonymous and known, based purely on their movement behavior. We validate our approach using publicly available flight data, comprising several hundred thousand flights with tens of millions of states in conjunction with meta information obtained from publicly available aircraft registries. Our results show that we can obtain the category of an aircraft with a likelihood of almost 90\%, based on features obtained from 30 flights or fewer. In cases where no metadata is publicly available for an aircraft, we show that our approach can be used to create this data, which is necessary for many research projects based on air traffic communication. Finally, we have examined a case study showing that it is possible to automatically discover sensitive aircraft in a large data set using Classi-Fly, including police, surveillance and military aircraft.

Future work in this area can focus on defining even more aircraft categories relevant for new and existing research applications. Specifically, we plan to extend the existing categories to include UAV and other non-standard aircraft such as gliders or ULAC. This requires these aircraft categories to have sufficiently broad equipage with ADS-B transponders or alternatives such as FLARM but can be a worthwhile expansion of coverage in many countries where FLARM is popular.\footnote{FLARM is a cooperative low-cost collision avoidance system developed for the gliding community \cite{santel2014glider}; its signals are collected by some web trackers.} Differences in behaviour across institutions and global regions could also be of much further interest.


\bibliographystyle{ACM-Reference-Format}
\bibliography{tist}

\end{document}